\documentclass[runningheads,hidelinks]{llncs}
\usepackage{graphicx}   
\usepackage[T1]{fontenc}
\usepackage{graphicx}
\usepackage{amsmath}
\usepackage{tabularx}
\usepackage{hyperref}
\usepackage{multirow}
\usepackage{amssymb}
\usepackage{xcolor}

\usepackage{orcidlink}
\usepackage[misc]{ifsym}

\begin{document}
\title{Ontology-Enhanced Knowledge Graph Completion using Large Language Models}

\titlerunning{Ontology-Enhanced KGC using LLMs}

\author{
Wenbin Guo\inst{1}\orcidlink{0009-0006-1649-4131} \and
Xin Wang\inst{1}\orcidlink{0000-0001-9651-0651}\textsuperscript{(\Letter)} \and
Jiaoyan Chen\inst{2}\orcidlink{0000-0003-4643-6750} \textsuperscript{(\Letter)} \and
Zhao Li\inst{1}\orcidlink{0000-0003-3996-255X} \and
Zirui Chen\inst{1}\orcidlink{0000-0002-7916-6745}
}
\authorrunning{W. Guo et al.}
%
\institute{Tianjin University, College of Intelligence and Computing, Tianjin 300350, China\\
\email{\{wenff, wangx, lizh, zrchen\}@tju.edu.cn} \and
University of Manchester, Department of Computer Science, Manchester M13 9PL, United Kingdom\\
\email{jiaoyan.chen@manchester.ac.uk}}


\maketitle          
\begin{abstract}
    Large Language Models (LLMs) have been extensively adopted in Knowledge Graph Completion (KGC), showcasing significant research advancements. However, as black-box models driven by deep neural architectures, current LLM-based KGC methods rely on implicit knowledge representation with the concurrent dissemination of erroneous knowledge, thereby hindering their ability to produce confirmative reasoning outcomes. 
    We aim to integrate neural-perceptual structural information with ontological knowledge, leveraging the powerful capabilities of LLMs to achieve a deeper understanding of the intrinsic logic of the knowledge. We propose an \textbf{O}ntology-enhanced \textbf{L}LM-based \textbf{KGC} method --- OL-KGC.
    It first leverages neural perceptual mechanisms to effectively embed structural information into the textual space, and then uses an automated extraction algorithm to retrieve ontological knowledge from the knowledge graphs (KGs) that needs to be completed, which is further transformed into a textual format that can be processed by LLMs to enhance their logical capability.
    We conducted extensive experiments on three widely-used benchmarks --- FB15K-237, UMLS, and WN18RR. The experimental results demonstrate that OL-KGC significantly outperforms existing mainstream KGC methods across multiple evaluation metrics, achieving state-of-the-art performance. The implementation of the algorithm and related data have been open-sourced\footnote{\url{https://github.com/xiumu-gg/OL-KGC}}.

\end{abstract}
\section{Introduction}

Knowledge Graphs (KGs) \cite{KG} represent a foundational paradigm for structured knowledge representation, serving as a cornerstone in the progression from perceptual intelligence to cognitive intelligence. KGs underpin numerous critical applications across diverse domains, including information retrieval \cite{KG-RAG}, question answering \cite{beyond}, and recommendation \cite{learning}.
KGs often organize real-world knowledge through triples in form of (head entity, relation, tail entity), enabling the structured representation of relational facts. Their expressive power and reasoning capabilities are further augmented by the incorporation of ontological frameworks and symbolic logic reasoning methods.
However, KGs often suffer from incompleteness due to various reasons. For example, algorithms designed to extract structured information from data often struggle to comprehensively capture the full spectrum of information in large datasets.
The Knowledge Graph Completion (KGC) is to address this challenge by leveraging existing knowledge and structural patterns to (iteratively) infer missing knowledge \cite{embedding}.

\begin{figure}[!t]
\centering
\includegraphics[width=\linewidth]{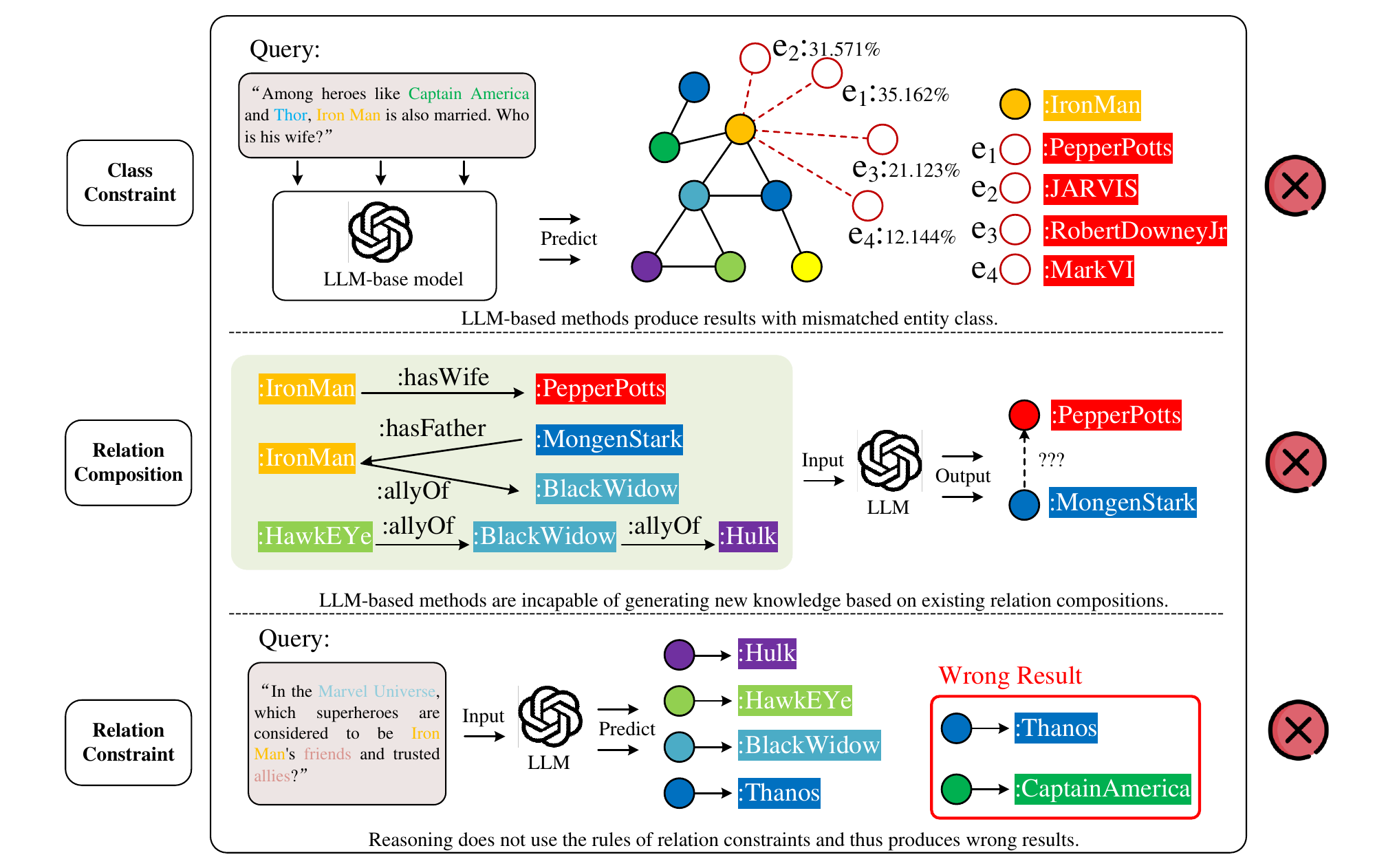}
\caption{LLM-based methods that fail to effectively utilize ontological knowledge.}
\label{fig:motivation}
\end{figure}

Due to the powerful capabilities of Large Language Models (LLMs), the field of KGC has undergone a significant transformation.
Recent LLM-based methods to KGC leverage the advanced contextual and semantic understanding capabilities of LLMs to enhance the precision and completeness of extracted entity and relation information, thereby improving the overall quality and reliability of results\cite{LLMsummary2025}.
These methods can be broadly categorized into two approaches: 1) leveraging existing paradigms such as zero-shot reasoning \cite{LLM-related} and instruction tuning \cite{LLM1} to translate structured KG predictions into LLM-based text predictions; and 2) integrating non-textual structural information from KGs into LLMs through correspondences between KG embeddings and LLM tokens. 
While the methods are capable of handling complex nonlinear relations, the absence of a mechanism for factual verification does not ensure the veracity and reliability of the generated content \cite{LLM3}.

As depicted in Figure \ref{fig:motivation}, LLM-based methods often fail to effectively leverage ontological information when verifying the accuracy of test triples, primarily in the following three aspects. First, these methods exhibit a lack of sensitivity to entity classes, and are unable to fully utilize the constraints imposed by relation range and domain to filter out the interference from erroneous triples. Second, they are incapable of performing reasoning based on existing relational paths within the KGs, thereby failing to accurately identify newly generated triples through relation compositions. Third, these methods cannot apply complex ontological knowledge, such as equivalence and disjointness relations, and are prone to deriving conclusions that are completely contrary to the facts during the reasoning process.
The underlying cause of this phenomenon is that LLMs do not incorporate an explicit symbolic reasoning mechanism during the prediction process. Instead, they rely excessively on the semantic correlation between entities, usually lacking fundamental logical reasoning capabilities. Although previous studies \cite{KGR2024} have attempted to address this issue by injecting rules into KG embeddings and constraining the quality of rule embeddings through loss functions, the results have been unsatisfactory.
Therefore, how to utilize the extensive knowledge domain covered by LLMs and their ability to capture complex data patterns to achieve a deep integration of vector space and symbolic space is the core issue of this research.

\begin{figure}[!t]
\centering
\includegraphics[width=\linewidth]{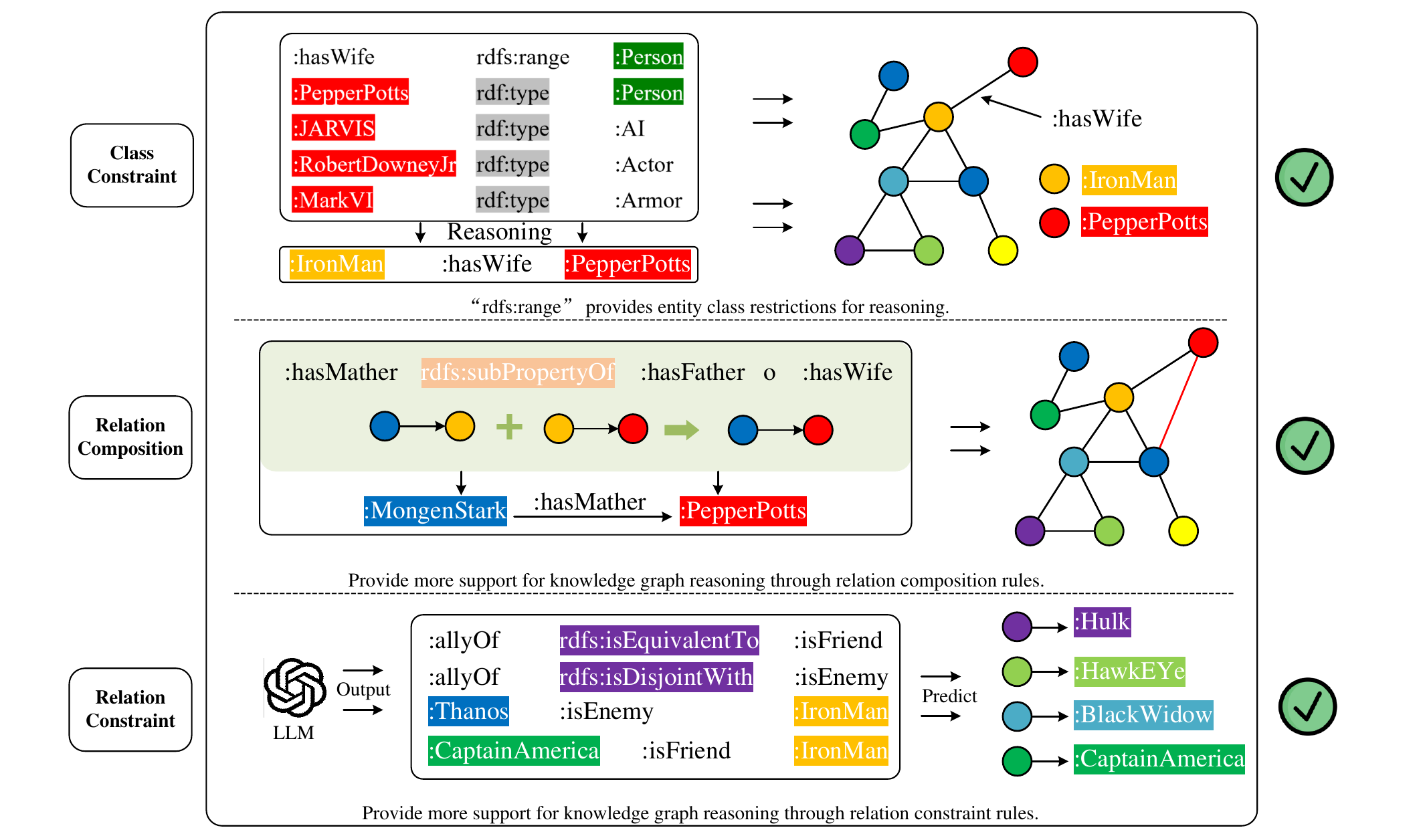}
\caption{The role of ontological knowledge in reasoning.}
\label{fig:motivation2}
\end{figure}

Given that a KG for completion is often short of ontological knowledge, this study first proposes an LLM-based automatic extraction algorithm to derive ontological knowledge from the KG that needs to be completed.
As shown in Figure \ref{fig:motivation2}, these extracted ontological knowledge cover the following key information: 
\begin{itemize}
    \item The range and domain of each relation, which can effectively validate triples according to the classes of their entities.
    \item The relation compositions based on existing relational paths, which can enabling the inference of new knowledge.
    \item Complex relation constraints beyond relation domain and range, such as relation equivalence and disjointness. 
\end{itemize}

We rewrite the extracted ontological knowledge, which is originally represented as RDF triples, into text so as to ensure that the LLM can fully comprehend their semantics. Meanwhile, we align the neural KG embeddings in the vector space with the text space tokens in the LLM, and incorporate the symbolic KG structural information and ontological knowledge into the LLM based on its tuning, thereby achieving effective neural-symbolic integration. {To the best of our knowledge, this work is the first to systematically integrate explicit ontological knowledge --- encompassing class constraints and relation compositions --- into LLMs through the joint exploitation of textual inference and structured symbolic logic.} Experimental results demonstrate that OL-KGC achieves SOTA performance in the field of KGC. In addition, we conducted ablation studies to individually assess the contributions of KG structural information and ontological knowledge. Furthermore, we assessed the influence of various types of ontological knowledge on reasoning performance within the model.

The main contributions of this paper can be summarized as follows:

\begin{itemize}
    \item We propose a novel method for integrating neural KG embeddings, and symbolic KG structural information and ontological knowledge based on LLMs and their tuning, achieving effective neural-symbolic integration and significantly enhancing LLM reasoning for KGC.
    \item We design an LLM-based automatic ontology extraction algorithm that can automate the ontology framework design for general KGs. This approach improves the semantic richness and the consistency and coherence of the relations and overall organization within KGs.
    \item Through extensive experiments, we demonstrate the performance superiority of OL-KGC in the KGC task. Additionally, we release all related code and data to provide a reproducible benchmark for future research.
\end{itemize}

\section{Related Works}

\subsection{Traditional Embedding-based KGC Models}

Traditional KGC models often adopt geometric embedding, assuming relational consistency, where relations imply a close alignment of entity vectors through specific operations \cite{MSHE}. Models enhance embedding quality using various operator spaces \cite{KGEditor}, such as TransE \cite{TransE}, which treats relations as translations between entity vectors. Subsequent studies have leveraged advanced natural language processing techniques to model complex relational patterns, including hyperplanes \cite{TransH}, Riemannian spheres \cite{5E}, relational mapping spaces \cite{TransD}, multi-mapping spaces \cite{TransM}, rotational spaces \cite{RotatE}, and convolutional spaces \cite{ConvE,ConvKB,ConvR}.

The introduction of attention mechanisms has enhanced the flexibility and efficiency of KGC models in deep learning \cite{CSANN}. RelAtt refines embeddings using a relation-aware attention mechanism \cite{RelAtt}, HRAN aggregates features along relational paths \cite{HRAN}, and KGT5 and HittER leverage Transformers for scalable and context-aware representation learning \cite{KGT5,HittER}. Despite their success, embedding-based models still struggle with distinguishing semantically similar but class-inconsistent entities \cite{summar20241}.

\subsection{PLM-based and LLM-based KGC Models}

PLM-based KGC models leverage the implicit knowledge within Pre-trained Language Models (PLMs) which are usually encoder-based and the structured knowledge in KGs to predict missing triples \cite{KnowC}. KG-BERT is a pioneering work in this domain \cite{KG-BERT}, which takes the textual information of triples as input and predicts the plausibility of triples through fine-tuning the BERT model. However, integrating the implicit knowledge of PLMs with KGs has always been a key challenge for such methods.

In contrast to leveraging contextual embeddings from PLMs to represent entities and relations in KGs for predicting missing triples, LLM-based KGC models exploit the more powerful and flexible text-generation capabilities of LLMs to directly generate missing triples or provide more comprehensive reasoning over KGs by generating textual explanations or intermediate steps. Recently, LLM-based KGC methods have been primarily categorized into two types \cite{LLM2}:

The first approach leverages paradigms like zero-shot reasoning and instruction tuning to translate structured KG information into text-based predictions via LLMs \cite{LLM3}. For example, MuKDC uses LLMs to generate subgraph-based questions for KGC \cite{MuKDC-1}, while KLR-KGC enhances LLM reasoning by retrieving analogical and subgraph knowledge from KGs \cite{KLR-KGC}.

The second approach incorporates structured KG information into LLMs by aligning KG embeddings with LLM tokens, improving their understanding and representation of KGs. TEA-GLM employs LLMs as zero-shot learners for graph machine learning tasks \cite{TEA-GLM}. KGEditor converts paths into Chain-of-Thought prompts for multi-hop link prediction \cite{KGEditor}. KoPA, the current state-of-the-art, introduces a knowledge prefix adapter to align structural embeddings with LLM tokens, effectively integrating KG information into LLMs \cite{KoPA}.

\subsection{Neuro-Symbolic Integration for KGC}

Besides embedding and language model-based methods, there are some neuro-symbolic methods for KGC that focus on the integration of symbolic reasoning and neural networks for enhancing reasoning capabilities and interpretability, often with mutual feedback between symbolic and neural components.

Neuro-symbolic approaches leverage logical rules to expand reliable data and enhance embedding quality \cite{TransO}. NeoEA aligns entity embeddings with ontological knowledge to reduce distribution discrepancies \cite{NeoEA}, while SymRITa employs symbolic rule graphs and a logical transformer to address predicate non-commutativity \cite{SymRITa}. Other methods enforce logical constraints via loss functions, such as IterE, which integrates rules into embeddings \cite{IterE}. GCR uses graph convolution networks with rule-enhanced embeddings \cite{GCR}, and KALE integrates embeddings and logical rules through a loss function \cite{KALE}.

However, these methods compromise the precision and interpretability of symbolic reasoning, facing challenges in integrating vector and symbolic spaces. The emergence of LLMs, which encapsulate extensive knowledge and complex patterns, presents new opportunities for this integration \cite{FLRS}.

\section{Notations and Preliminaries}

A KG is represented as $\mathcal{G} = \{ \mathcal{E}, \mathcal{R}, \mathcal{T}, \mathcal{H}, \mathcal{D} \}$, where $\mathcal{E}$ and $\mathcal{R}$ denote the sets of entities and relations, respectively, and $\mathcal{T} = \{(h, r, t) \mid h, t \in \mathcal{E}, r \in \mathcal{R} \}$ represents the set of triples. 
$\mathcal{H}$ refers to the ontological information, represented as a set of RDF triples.
The ontological information we consider includes the following types:

\begin{itemize}
    \item \textbf{Relation Domain and Range Constraints:} For a given entity $e \in \mathcal{E}$, the entity class constraint stipulates that $e$ belongs to a specific class $a$. Furthermore, if $e$ also belongs to a superclass class of $a_s$, it is denoted as a $a_s \subseteq a$, indicating that $a$ inherits the attributes of $a_s$. The properties \texttt{:range} and \texttt{:domain} can be utilized to restrict the classes of the head entity $h$ and the tail entity $t$ of a specific relation $r$. For instance, for the relation \texttt{:hasWife}, the \texttt{:range} and \texttt{:domain} property can restrict the head entity $h$ to the class \texttt{:Person} and the tail entity $t$ to the class \texttt{:Person}. Given that the entity \texttt{:PepperPotts} belongs to the class \texttt{:Women}, which is a subclass of \texttt{:Person}, the entity \texttt{:PepperPotts} complies with the \texttt{:range} restriction of the relation \texttt{:hasWife}, thereby validating the \texttt{:range} constraint.
    
    \item \textbf{Relation Compositions:} Relation compositions enable the inference of new relations within a $\mathcal{G}$. For instance, given two RDF triples \\(\texttt{:MongenStark} \texttt{:hasFather} \texttt{:IronMan}) and (\texttt{:IronMan} \texttt{:hasWife} \\
    \texttt{:PepperPotts}), a new relation can be inferred through the relation composition (\texttt{:hasMother} \texttt{rdfs:subPropertyOf} \texttt{:hasFather} \texttt{o} \texttt{:hasWife}). This results in the new inference (\texttt{:MongenStark} \texttt{:hasMother} \texttt{:PepperPotts}), thereby deducing the previously unknown fact that PepperPotts is the mother of MongenStark based on the existing knowledge in the $\mathcal{G}$.
    
    \item \textbf{Complex Relation Constraints:} Complex relations, such as equivalence and disjointness relation, are employed to constrain the results of inference. For equivalence relations, the $\mathcal{G}$ is expanded. For example, given that \texttt{:allyOf} and \texttt{:isFriend} are equivalent relations, the triple (\texttt{:IronMan} \\ \texttt{:allyOf} \texttt{:CaptainAmerica}) can be expanded to yield the new triple \\ (\texttt{:IronMan} \texttt{:isFriend} \texttt{:CaptainAmerica}). For disjointness relations, the inference results are validated. For instance, given that \texttt{:isEnemy} and \texttt{:isFriend} are disjointness relation, if the triple (\texttt{:IronMan} \texttt{:isEnemy} \texttt{:Thanos}) is true, then the triple (\texttt{:IronMan} \texttt{:isFriend} \texttt{:Thanos}) must be false.
\end{itemize}
 
$\mathcal{D}$ denotes the descriptive information of each entity and relation, where each entity and relation is uniquely identified in the real data. Specifically, we use $\mathcal{D}(e)$ and $\mathcal{D}(r)$ to represent the textual descriptions of each entity and relation, respectively. For example, in the widely used dataset FB15K-237, we have $\mathcal{D} (\text{/m/0dzlbx}) = \text{``Iron Man''} $.
  
We introduce an LLM $\mathcal{M}$ as the text decoder for KGC. The input sequence $\mathcal{S}$ consists of three components: the instruction prompt $\mathcal{I}$, the triple prompt $\mathcal{X}$, and the ontology prompt $\mathcal{O}$. 
$\mathcal{I}$ provides manual guidelines for completing the task, such as the instruction ``Please help determine whether the triple (h,r,t) is a valid triple.'' Meanwhile, $\mathcal{X}$ contains textual descriptions of the triple symbols, represented as $\mathcal{X} (h, r, t) = \mathcal{D} (h) \oplus \mathcal{D} (r) \oplus \mathcal{D} (t) $ , where $(h, r, t)$ is the triple, and $\oplus$ denotes the concatenation. $\mathcal{O}$ represents the textual form of the ontological information, which is derived by transforming a relevant part of $\mathcal{H}$ into corresponding text.

We focus on triple classification which is a fundamental task of KGC. Following most LLM-based methods that reformulate tasks as text generation problems, we expect our LLM model $\mathcal{M}$, given a text sequence $\mathcal{S} = \mathcal{I} \oplus \mathcal{X} \oplus \mathcal{O}$ of an input triple, outputs a response of either ``True'' or ``False'' for validating this triple. Unlike text classification, the complex semantics of KGs cannot be captured by a single description. Without KG structural information, LLMs may primarily encapsulate isolated term-level knowledge, which restricts their capacity for comprehensive understanding. Despite their vast general knowledge, LLMs are prone to hallucinations due to insensitivity to fine-grained facts. Thus, effective KGC with LLMs requires vector-based methods to provide structured knowledge, enabling structure-aware reasoning.

\section{Methodology}

\begin{figure*}[!t]
\centering
\includegraphics[width=\linewidth]{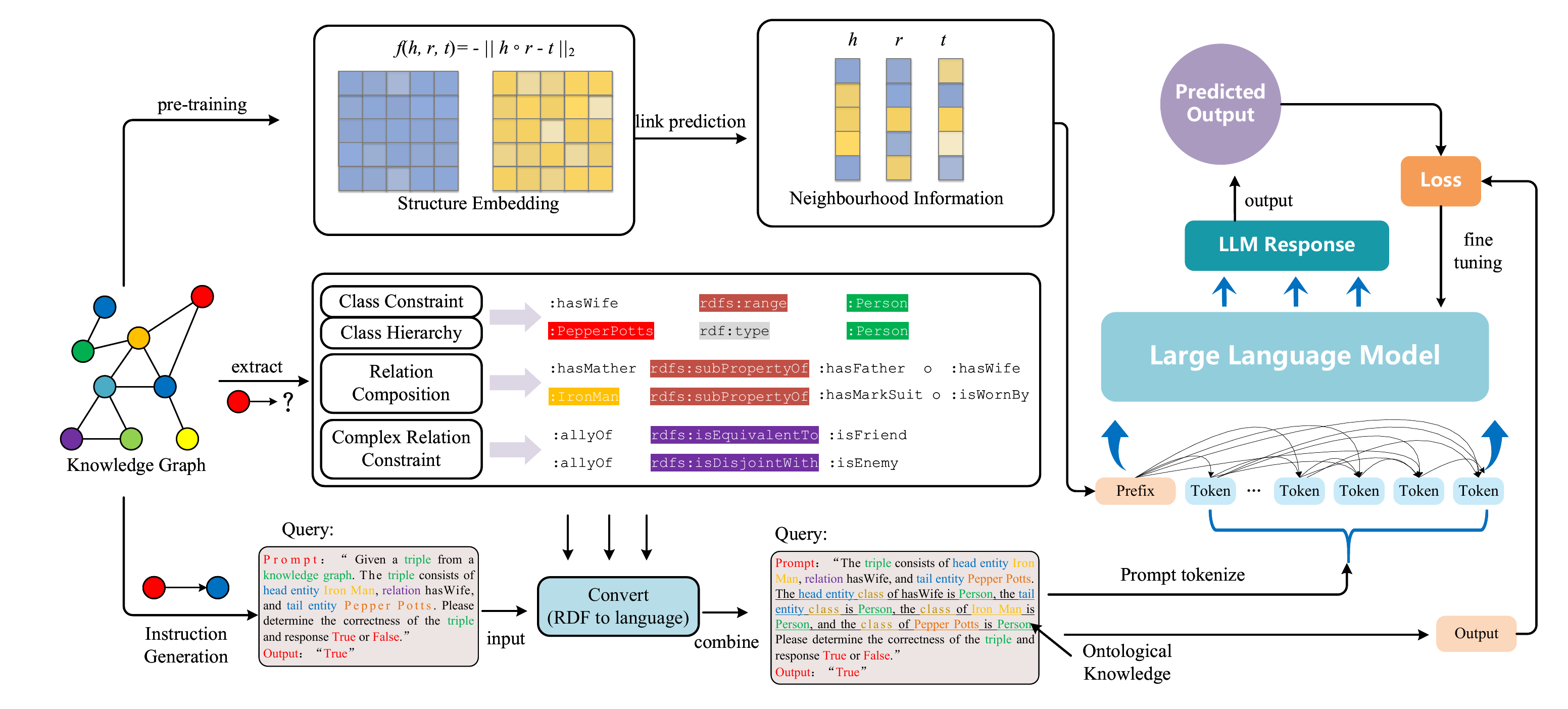}
\caption{The overall framework of OL-KGC.}
\label{fig:framework}
\end{figure*}

    This section elucidates how OL-KGC integrates LLMs with vectorized structural and ontological knowledge. Initially, ontological knowledge is extracted in RDF format and transformed into text to serve as part of the input for the LLM. Subsequently, traditional knowledge graph embedding (KGE) algorithms are employed to extract structural information from the KG, which is then provided to the LLM in prefix form to enhance its understanding of the complex structure of the KG. Finally, the LLM integrates the vectorized structural and ontological knowledge, and the completion performance is optimized through fine-tuning using the LoRA method. The framework of OL-KGC is shown in Figure \ref{fig:framework}. 

\subsection{Ontological Knowledge Integration} 

KGs typically store knowledge in symbolic form. For instance, in the FB15K-237 dataset, entities are represented by symbols such as /m/0dzlbx, which are not directly suitable for extracting ontological information. To address this, we uesd an automated script to map the symbols to the Freebase database, generating textual descriptions for each entity. For example, the entity /m/0dzlbx is mapped to $\mathcal{D} (\text{/m/0dzlbx}) = \text{``Iron Man''} $.

We further leverage the powerful generative capabilities of LLMs to automatically derive corresponding classes for each entity, generate equivalence, disjointness, and composition rules for each relation, and create class inheritance relationships for each class. Based on the output of the LLM, we designed an algorithm to automatically extract ontological knowledge represented in RDF format, denoted as $\mathcal{H}$, from its generated results. To ensure the accuracy of the extracted ontological knowledge $\mathcal{H}$, we introduced a mechanism for manual expert verification.

In addition, we constructed an RDF language mapping dictionary. For a given test triple $(h,r,t)$, we retrieve relevant RDF triples from $\mathcal{H}$ and transform them into natural language text using predefined templates. For example, the RDF triple (\texttt{:IronMan} \texttt{rdf:type} \texttt{:Person}) is mapped to ``The class of entity \texttt{:IronMan} is \texttt{:Person}.'' This mapping method helps the LLM understand the logical structure of the KG. The process can be represented as follows:

\begin{equation}
    \mathcal{O} = \mathcal{B} \left( \sum_{\mathcal{H}_i \in \mathcal{H}} \mathcal{S}(\mathcal{H}_i, (h, r, t)) \right)
\end{equation}
where $\mathcal{S( \cdot)}$ is an indicator function that determines the relevance between $\mathcal{H}_i$ and the test triple by checking whether there exist entity, relation, or corresponding entity class related to the test triple within $\mathcal{H}_i$. When $\mathcal{H}_i$ is relevant to the test triple $(h,r,t)$, $\mathcal{S( \cdot)}$ returns $\mathcal{H}_i$; otherwise, it returns nothing.
$\mathcal{B}$ maps RDF triples to natural language text via a predefined dictionary.
$\mathcal{O}$ is the textual description of ontological knowledge of the KG and will serve as part of the input to the LLM.

\subsection{Structural Information Integration}

$\mathcal{G}$ contains rich structural information, but its complexity makes it difficult to represent through explicit reasoning processes. To capture the intricate structural information within the KG, we utilize self-supervised embedding pretraining. For each entity $e \in \mathcal{E}$ and each relation $r \in \mathcal{R}$, we learn embeddings $\mathbf{e} \in \mathbb{R}^{d_e}$ and $\mathbf{r} \in \mathbb{R}^{d_r}$, where $d_e$ and $d_r$ represent the dimensions of the entity and relation embeddings, respectively, to represent neighborhood structure information. We evaluated existing KGC models experimentally, and the results show that using a rotational operator most effectively extracts the structural features of the KG. The scoring function is defined as:  
\begin{equation}
    f(h, r, t) = -\mid \mathbf{h} \circ \mathbf{r} - \mathbf{t} \mid_2
\end{equation}
 where $\circ$ denotes the Hadamard product between the head entity embedding and the relation embedding, and $\mid \cdot \mid_2$ represents the $l_2$-norm of the complex vector. By improving the quality of embeddings $\mathbf{e}$ and $\mathbf{r}$, we are able to extract the complex structural information of the KG. As the score of the positive triples increases, the entity embeddings $\mathbf{e}$ and relation embeddings $\mathbf{r}$ are optimized, thus capturing subgraph structures and relation patterns within the KG. This approach has been demonstrated to be effective in several embedding-based KGC methods \cite{RotatE}, which can capture classical structural information such as relation patterns and distributed entity representations.

 As shown in Figure \ref{fig:framework}, for the test triple $(h,r,t)$, we extract its corresponding vector representation through structural embedding. The dimensionality of this vector is consistent with that of the token vectors in the LLM. To enhance the perception of the structural information in KG, we innovatively concatenate the vectors corresponding to $(h,r,t)$ and transform the structural embedding from the vector space to the LLM token text space via a linear adapter. This transformed representation is then used as a prefix for the input to the LLM. 

 \begin{equation}
    \mathbf{v}_{(h,r,t)} = \mathbf{W}_{\text{linear}} (\mathbf{v}_h \oplus \mathbf{v}_r \oplus \mathbf{v}_t) + \mathbf{b}_{\text{linear}}
\end{equation}
there $\oplus$ denotes the concatenation operation, $\mathbf{W}_{\text{linear}}$ is the weight matrix of the linear layer, and $\mathbf{b}_{\text{linear}}$ is the bias vector. The resulting vector $\mathbf{v}_{(h,r,t)}$ serves as the prefix for the input to the LLM. This design enables the structural information to directly influence the generation process of each subsequent token, thereby providing richer semantic and structural context for the KGC. During training, the pre-trained structural embeddings are frozen, while the linear adapter is optimized to learn the mapping from structural knowledge to textual representation. This mapping can then be generalized to new triples during the inference stage.

\subsection{LLM Fine Tuning}

The powerful reasoning capabilities of LLMs provide a more effective tool for integrating vectorized KG structural information and ontological information. The input to the LLM includes the instruction, the textual interpretation of the triple, and the ontological information, which can be represented as follows:

\begin{equation}
    \mathcal{S} = \mathcal{I} \oplus \mathcal{X} \oplus \mathcal{O}
\end{equation}
there $\mathcal{X}$ denotes the descriptive text of the triple, and $\mathcal{O}$ represents the ontological information that describes the logic within the KG. We concatenate these two types of textual information with the instruction $\mathcal{I}$ to form the input to the LLM. Additionally, the structural embedding is transformed from the vector space to the LLM token text space and serves as a prefix during the LLM's generation phase, providing structural information of the KG for each token generated in this phase.

\begin{equation}
    \mathcal{A}_{\text{token}} = \mathbf{v}_{(h,r,t)} \oplus \text{tokenize}( \mathcal{S})
\end{equation}
there $ \mathcal{A}_{\text{token}}$ represents the complete token sequence input to the LLM, and the $\text{tokenize}(\cdot)$ function denotes the process of tokenizing the text. Through uni-directional attention decoding, the LLM can perceive the structural information of the KG while generating each token, which is beneficial for prediction.

To train the large model to better predict the triples, thereby enhancing its performance in KGC, we fine-tune the model using LoRA by minimizing the following loss function:

\begin{equation}
    \mathcal{L} = \gamma \cdot \mathcal{D}(\mathcal{M} (\mathcal{A}_{\text{token}}), Output)
\end{equation}
where $\gamma$ is a weighting factor used to adjust the intensity of the loss; $\mathcal{D}$ is a discrepancy measurement function that quantifies the difference between the output and the expected result. The model parameters are optimized through backpropagation based on this discrepancy, which helps the LLM perform more accurate KGC.

\section{Experiments}

We conducted experiments to verify the superiority of OL-KGC. The experiments were guided by answering the following four questions: \textbf{Q1:} What are the advantages of this method compared with various traditional KGC methods? \textbf{Q2:} What roles do the vectorized structural information and ontological knowledge play respectively? \textbf{Q3:} Are the ontological knowledge effectively incorporated into the LLM, and what is the impact of the ontological knowledge?

\subsection{Experimental Settings}

\begin{table}
\centering
\caption{Dataset Statistics.}\label{tab:sets}
\begin{tabular}{c|c|c|c|c|c|c|c}
\hline
\textbf{Dataset} & \textbf{Rel} & \textbf{Ent} & \textbf{Tri} & \textbf{Ont} & \textbf{Ont-cla} & \textbf{Ont-rel} & \textbf{Ont-com}\\
\hline
UMLS-O     & 46 & 135 & 6,529 & 165 & 106 & 13 & 46 \\
FB15K-237O & 237 & 14,541 & 310,116 & 18,763 & 16,073 & 93 & 2597 \\
MN18RR-O  & 11  &  40,943 & 94,003 & 4,710 & 4,683 & 9 & 18  \\
\hline
\end{tabular}
\end{table}

\textbf{Datasets}.
We conducted experiments using three public datasets: UMLS, FB15K-237, and MN18RR. 
Based on the fundamental datasets, we introduced ontological information in the form of RDF triples to each dataset through the LLM, which encompassed rules such as entity classes, relation composition, equivalence relations, and disjointness relations. 
This resulted in the creation of the UMLS-O, FB15K-237O, and MN18RR-O datasets. 
The fundamental datasets are composed in the form of symbols. For the FB15K-237 and WN18RR datasets, we extracted relevant entity description text information from Freebase and WordNet respectively. As for the UMLS dataset, we introduced the Metathesaurus ontology\footnote{\url{https://www.nlm.nih.gov/research/umls/knowledge_sources/metathesaurus}} to obtain relevant information.
The enriched datasets will be publicly available\footnote{\url{https://github.com/xiumu-gg/OL-KGC/data}}. Detailed information is provided in Table \ref{tab:sets}. 
The symbols Rel, Ent, Tri, Ont represent the quantities of relations, entities, triples, and ontological information, respectively. The ontological information is further subdivided into three distinct categories: Ont-cla, which pertains to entity classes; Ont-rel, which encompasses relation composition; and Ont-com, which represents constraints on complex relations.

{For the data leakage scenario where the LLM may have directly seen or learned from the data used during testing, we have checked that our datasets FB15K-237 and WN18RR, as well as the original KGs they come from, are not used in any training stage of the GLM model which used in our experiments. LLM may have seen text with similar semantics as the structured triples, but such memorization is a key feature of LLM itself. We intentionally selected UMLS --- a domain-specific biomedical dataset --- to reduce the overlap with the pretraining corpora.}

\begin{table*}
\centering
\caption{The experimental results for triple classification are evaluated using accuracy (Acc), precision (P), and F1-score (F1). The best results are marked in \textbf{bold}, and the second-best results are underlined. The performance metrics for the models are primarily sourced from \cite{KoPA}. For models without reported results in \cite{KoPA}, we conducted local experiments to obtain the corresponding values. In cases where certain results could not be reproduced, a `-' symbol is used to indicate missing data.}
\label{tab:result}
\begin{tabularx}{\textwidth}{c@{\hspace{10pt}}|c@{\hspace{7pt}}c@{\hspace{7pt}}c@{\hspace{7pt}}|c@{\hspace{7pt}}c@{\hspace{7pt}}c@{\hspace{7pt}}|c@{\hspace{7pt}}c@{\hspace{7pt}}c}
\hline
\textbf{Model} & \multicolumn{3}{c|}{\textbf{UMLS-O}} & \multicolumn{3}{c|}{\textbf{FB15K-237O}} & \multicolumn{3}{c}{\textbf{WN18RR-O}} \\
\cline{2-10}
& \textbf{Acc} & \textbf{P} & \textbf{F1} & \textbf{Acc} & \textbf{P} & \textbf{F1} & \textbf{Acc} & \textbf{P} & \textbf{F1} \\
\hline
TransE & 84.46 & 86.13 & 83.82 & 69.85 & 70.80 & 68.75 & 62.55 & 51.23 & 52.87 \\
ComplEx & 90.76 & 89.90 & 90.85 & 65.68 & 66.45 & 64.85 & 77.43 & 71.83 & 69.91 \\
RotatE & 91.95 & 90.14 & 92.15 & 68.82 & 69.43 & 67.10 & 78.61 & 74.82 & 70.06 \\
ConvE & 90.08 & 88.34 & 90.26 & 68.59 & 69.91 & 66.88 & 77.05 & 71.67 & 69.50 \\
\hline
IterE & 80.37 & 83.29 & 81.89 & 65.73 & 66.62 & 65.10 & 71.85 & 67.81 & 65.06 \\
NeoEA & 85.62 & 84.62 & 83.47 & 66.92 & 67.29 & 66.38 & 72.61 & 69.42 & 67.68 \\
\hline
KG-BERT & 77.25 & 70.84 & 80.24 & 55.83 & 53.42 & 67.71 & 58.80 & 57.28 & 64.38 \\
KnowC & 80.84 & 86.97 & 84.89 & 59.72 & 60.12 & 66.83 & 68.16 & 64.80 & 66.57 \\
\hline
KG-LLaMA & 85.68 & 87.77 & 85.30 & 74.77 & 67.40 & 79.27 & \underline{80.85} & 76.83 & \textbf{80.54} \\
KoPA & 92.58 & \underline{90.85} & \underline{92.70} & \underline{77.65} & 70.81 & \underline{80.81} & - & - & - \\
MuKDC & 89.13 & 90.07 & 86.85 & 72.70 & \underline{75.92} & 74.63 & 77.55 & 77.83 & 73.82 \\
KGEditor & \underline{92.85} & 86.96 & 90.56 & 63.47 & 68.73 & 70.03 & 80.63 & \underline{81.22} & 77.81 \\
\hline
\multicolumn{1}{c|}{\textbf{OL-KGC}} & \textbf{93.10} & \textbf{92.32} & \textbf{93.16} & \textbf{80.41} & \textbf{76.26} & \textbf{84.66} & \textbf{81.45} & \textbf{83.50} & \underline{78.65} \\
\hline
\end{tabularx}
\end{table*}

\textbf{Baseline Methods}.
In the experimental section, we perform a comparative analysis across four categories of KGC models: traditional KGE models, neuro-symbolic integrated KGC models, PLM-based KGC models, and LLM-based KGC models. For traditional KGE models, we selected four baseline models: TransE \cite{TransE}, ComplEx \cite{ComplEx}, RotatE \cite{RotatE}, and ConvE \cite{ConvE}, which represent different approaches to capturing semantic relations in KGs via various scoring functions. The neuro-symbolic models IterE \cite{IterE} and NeoEA \cite{NeoEA} were also included for comparison. For PLM-based KGC models, we chose KG-BERT \cite{KG-BERT} and KnowC \cite{KnowC}, which utilize the semantic power of PLM-based models to integrate graph structure and semantic information. Additionally, we selected high-performance LLM-based models such as KG-LLaMA \cite{KG-LLaMA}, KoPA \cite{KoPA}, MuKDC \cite{MuKDC-1}, and KGEditor \cite{KGEditor} as baselines.

\textbf{Evaluation Metrics}.
Unlike conventional link prediction tasks, LLMs are not inherently suited for predicting the global ranking of specific entities. Consequently, we selected triple classification as the evaluation task for KGC models, which serves as a crucial subtask within the KGC framework. {We assessed model performance with the standard suite of metrics—accuracy, precision, recall, and F1-score—defined as follows: Accuracy is defined as the ratio of correctly classified triples (both positive and negative) to the total number of evaluated triples. Precision is the ratio of correctly predicted positive triples to the total number of triples predicted as positive. Recall is the ratio of correctly predicted positive triples to the total number of ground-truth positive triples. F1 is the harmonic mean of precision and recall.}

\subsection{Link Prediction Results}

The experimental results for triple classification are shown in Table \ref{tab:result}. 
Given that the core objective of this study is to evaluate the prediction accuracy and precision of the model, and considering the limitations of page space, we did not present the results of recall in the link prediction experiment section. We believe that precision and the F1 score can more accurately reflect the value of the model. The complete experimental results report has been detailed in the supplementary material for reference by the readers.

OL-KGC achieved state-of-the-art performance on average across the three datasets, with respective performance improvements of 2.238\%, 1.361\%, and 2.246\% on the UMLS-O, FB15K-237O, and WN18RR-O datasets. This indicates that the KGC method combining vectorized structural information and ontological information using LLM can achieve more accurate predictions. It is worth noting that the model reached state-of-the-art levels in terms of accuracy across different datasets, demonstrating its capability to accurately complete missing information. However, the model exhibited poor performance in terms of the F1 metric on the WN18RR-O dataset. In KGs, certain classes contain a significantly larger number of entities compared to others. 
The ontological knowledge we designed is unable to effectively constrain the prediction process, which can easily lead to the neglect of minority positive triples, thereby resulting in a relatively lower F1 metric compared to other metrics.

Compared to traditional KGE methods, LLM-based approaches achieve superior results. While PLM-based models possess complex structures, they fail to capture the structural information of triples, leading to poor performance on triple classification tasks. Traditional KGE methods are more suitable for reasoning on small datasets, and their reasoning results are more balanced across four different metrics, without the drawback of a significantly lower F1 score. LLM-based methods, on the other hand, are overall better suited for triple classification tasks, but exhibit significant disparities across different metrics.

On the small UMLS-O dataset, LLM-based methods do not exhibit overwhelming advantages, as smaller parameterized KGE models are capable of capturing sufficient structural and semantic information. However, on larger datasets, LLM-based approaches demonstrate more significant advantages. Specifically, the FB15K-237O dataset contains more relations, and OL-KGC is able to perform more accurate reasoning in KGs with a higher number of relations. In contrast, on the relatively simpler WN18RR-O dataset characterized by fewer relational patterns, OL-KGC lacks the capacity to infer richer relation-based ontological knowledge, leading to performance comparable to other current state-of-the-art LLM-based KGC methods (answering \textbf{RQ1}).

\begin{table}
\centering
\caption{The results of the ablation study on the FB15K-237O dataset are presented, with the best performance highlighted in \textbf{bold}.}
\label{tab:Neuro-Symbolic}
\begin{tabular}{c|c|c|c|c}
\hline
 & \textbf{Acc} & \textbf{P} & \textbf{R} & \textbf{F1} \\
\hline
w/o ontology & 67.81 & 61.41 & 69.08 & 69.40  \\
w/o structure & 74.35 & 70.81 & 79.38 & 77.44  \\
w/o structure \& ontology & 62.18 & 57.53 & 66.20 & 64.83  \\
\hline
\textbf{OL-KGC} & \textbf{80.41} & \textbf{76.26} & \textbf{86.44} & \textbf{84.66} \\
\hline
\end{tabular}
\end{table}

\subsection{Ablation Study}

\begin{figure}[!t]
\centering
\includegraphics[width=0.8\linewidth]{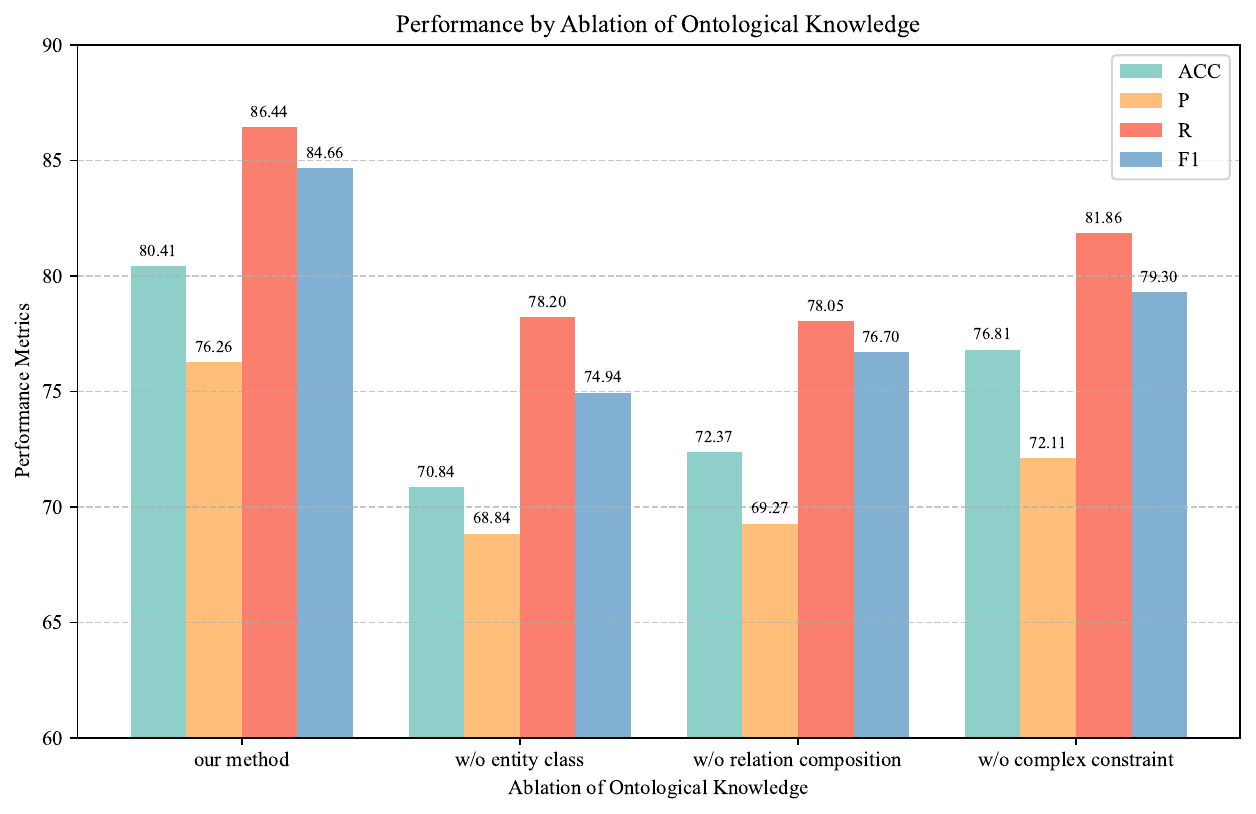}
\caption{The results of the ontology rule ablation experiments.}
\label{fig:ablation}
\end{figure}

\textbf{Structural Information and Ontological Information}.
The model achieved state-of-the-art performance by integrating vectorized structural information and ontological knowledge through the LLM. However, the specific contributions of the vectorized structural information and ontological knowledge to the final prediction results remain unknown. To investigate this, we conducted experiments by removing the ontological information from the input $\mathcal{S} = \mathcal{I} \oplus \mathcal{X} \oplus \mathcal{O} \oplus \mathcal{U}$ and eliminating the relevant text in the instructions to assess the impact of removing ontological knowledge (w/o ontology). We also randomized the prefixes of the neural-perceived structural information to cancel the implicit structural information influence (w/o structure). Additionally, we removed both structural information and ontological knowledge (w/o structure \& ontology) to evaluate the performance of a standard LLM on the triple classification task. The results on the FB15K-237O dataset are shown in Table \ref{tab:Neuro-Symbolic}.

he experimental results indicate that each module of OL-KGC contributes to the final performance of the model. The impact of structural information on the final outcome is relatively minor, ranging from 7.05\% to 8.50\%. The performance of a standard LLM on the triple classification task is significantly inferior, with a gap of 22.77\% to 24.48\% compared to OL-KGC. The influence of ontological knowledge on the model lies between 15.77\% and 20.08\%. Given that LLMs are inherently based on a black-box architectural design, the implicit knowledge provided by vectorized structural information serves as a supplement to LLMs. Ontological knowledge, on the other hand, offers logical connections for LLM reasoning. Therefore, it is reasonable that ontological knowledge has a more substantial impact on the results.

\textbf{Concrete Ontological Information}.
To investigate the impact of different classes of ontological information, we selectively removed distinct rules and assessed the effects of entity classes (w/o entity class), relation compositions (w/o relation composition), and complex constraints (w/o complex constraint) on the model. Specifically, for entity classes and class inheritance relations, we eliminated the RDF triples corresponding to the rules and the RDF triples that restrict the class of triples. For relation compositions, we removed all RDF rules related to relation compositions. Similarly, for complex relations, we deleted rules concerning equivalence, disjointness, and other complex relations. The results on the FB15K-237O dataset are illustrated in Figure \ref{fig:ablation}.

\begin{figure}[!t]
\centering
\includegraphics[width=\linewidth]{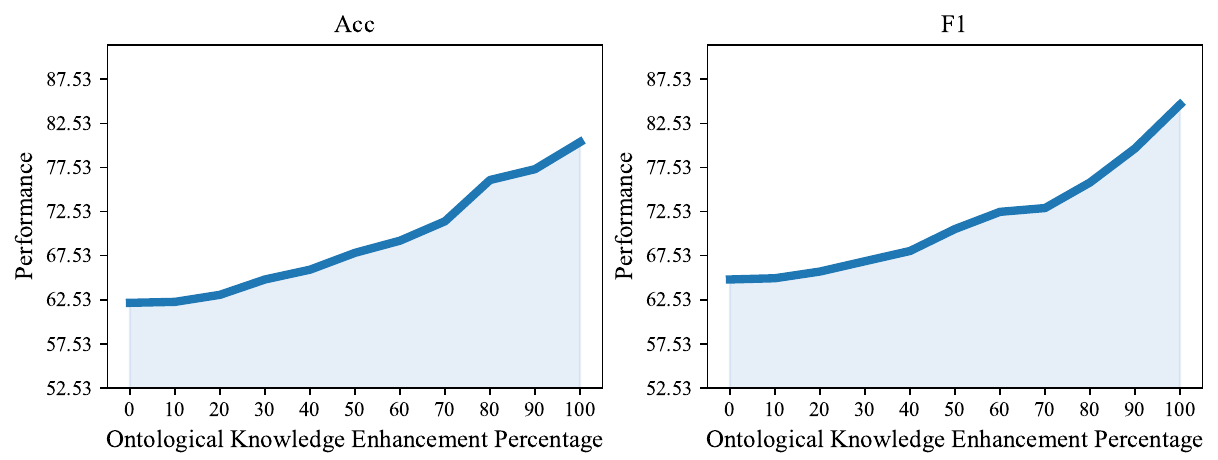}
\caption{The results of the ontological knowledge analysis experiment.}
\label{fig:Interpretability}
\end{figure}

{The experimental results indicate that OL-KGC mitigates hallucinations by injecting logical constraints derived from the underlying ontology, thereby enhancing inter-run consistency and improving inferential accuracy.} Specifically, the impact of entity class rules is relatively substantial, ranging from 9.53\% to 12.01\%. In contrast, the influence of complex relations on the model is relatively minor, falling between 4.60\% and 6.30\%. We attribute the differences to the fact that specialized complex category relations are not extensively present in the KG data; they primarily serve as supplementary information to aid the LLM in reasoning. Conversely, entity classes can impose constraints on every triple (answering \textbf{RQ2}).


\subsection{Ontological Knowledge Analysis Experiment}

We focus specifically on the impact of ontological knowledge on model performance. By gradually increasing the application of ontology rules, we compare the ACC and F1 under constrained and unconstrained conditions. This comparison visually demonstrates the positive impact of explicit rules on model performance, thereby validating the interpretability of the model in KGC. The results on the FB15K-237O dataset are shown in Figure \ref{fig:Interpretability}.

The dark-colored lines indicate the performance of the model after incorporating the corresponding percentage of ontological knowledge, while the shaded areas represent the performance gains attributable to the use of such knowledge—that is, correct reasoning outcomes derived from ontological analysis. Across all evaluation metrics, model performance consistently improves as the amount of ontological knowledge increases. This trend suggests that the incremental incorporation of ontological knowledge provides the model with progressively richer semantic constraints.

{Meanwhile, our approach is readily applicable to real-world KGs that embed rich ontological knowledge, with financial KGs serving as a paradigmatic example. Within these graphs, hierarchical concept taxonomies such as \texttt{:Enterprise} $\sqsupseteq$ \texttt{:SmallEnterprise} $\sqsupseteq$ \texttt{:HighRiskSmallEnterprise} enable OL-KGC to (i) infer class inheritance and (ii) verify constraint satisfaction, thereby safeguarding class-level consistency and enhancing the accuracy of downstream risk assessments} (answering \textbf{RQ3}).

\subsection{Case Analysis Experiment}

{To track the role of ontological knowledge, we conducted a case study, showcasing several correct and incorrect predictions from our method, with their explanation. In our experiments on the FB15K-237 dataset, we observed that the triple (\texttt{:LosAngeles}, \texttt{:locatedIn}, \texttt{:Country}) was incorrectly predicted as True by the LLM when no ontological knowledge was provided. This misjudgment arises from strong semantic associations between the entities, despite violating the range constraint of \texttt{:locatedIn} relation, which expects the tail entity to belong to \texttt{:Country} class (for example, entity \texttt{:America} satisfies this constraint), whereas the entity \texttt{:Country} itself does not belong to the \texttt{:Country} class. When ontological constraints --- specifically domain and range information --- were injected, OL-KGC correctly predicted this triple as False. Conversely, the valid triple (\texttt{:LosAngeles}, \texttt{:locatedIn}, \texttt{:Country}) was consistently predicted as True due to class consistency.}

\section{Conclusion}

This paper introduces OL-KGC, a KGC method that leverages LLMs. By integrating vectorized structural information and ontological knowledge, the performance and reasoning capability of the KGC are effectively enhanced. We introduced various types of ontological knowledge into existing KGs and utilized LLMs to incorporate symbolic ontological knowledge into reasoning. Experimental results demonstrate that OL-KGC achieves state-of-the-art performance in the field of KGC. 

\paragraph*{Supplemental Material Statement: We have published the algorithms for introducing ontological knowledge to existing KGs using LLM on GitHub \footnote{\url{https://github.com/xiumu-gg/OL-KGC/dataset-process}}. The datasets FB15K-237O, WN18RR-O, and UMLS-O, which were constructed on FB15K-237, WN18RR, and UMLS, are also open-sourced on GitHub \footnote{\url{https://github.com/xiumu-gg/OL-KGC/data}}. {The algorithmic implementations and complete experimental parameters --- including per-component computational costs and detailed hardware configurations --- are publicly available on GitHub}\footnote{\url{https://github.com/xiumu-gg/OL-KGC}}.}

\section*{Acknowledgments}
{This work was supported by the National Natural Science Foundation of China (62472311), the Key Research and Development Program of Ningxia Hui Autonomous Region (2023BEG02067), and the EPSRC Project OntoEm (EP/Y017706/1).}

\bibliographystyle{splncs04}
\bibliography{iswc}
%




\end{document}